\documentclass{article}
\usepackage{spconf,amsmath,graphicx,bigstrut, bm}
\usepackage{amsfonts,algorithm,algorithmic,caption,adjustbox, booktabs}
\usepackage{enumitem}

\setlist[itemize]{left=0pt}
\setlist[enumerate]{left=0pt}

\def\methname{LocalGCL }
\def\methnametrim{LocalGCL}

\title{LocalGCL: Local-aware Contrastive Learning for Graphs}

\name{Haojun Jiang$^1$, Jiawei Sun$^1$, Jie Li$^{1,2,}$, Chentao Wu$^1$} 
\address{$^1$Department of Computer Science and Engineering, Shanghai Jiao Tong University, China\\
$^2$MoE Key Lab of Artificial Intelligence, AI Institute,
Shanghai Jiao Tong University, China\\
}

\begin{document}
%
\maketitle
\begin{abstract}

Graph representation learning (GRL) makes  considerable progress recently, which encodes graphs with topological structures into low-dimensional embeddings.
Meanwhile, the time-consuming and costly process of annotating graph labels manually prompts the growth of self-supervised learning (SSL) techniques. As a dominant approach of SSL, Contrastive learning (CL) learns discriminative representations by differentiating between positive and negative samples. However, when applied to graph data, it overemphasizes global patterns while neglecting local structures. 
To tackle the above issue, we propose \underline{Local}-aware \underline{G}raph \underline{C}ontrastive \underline{L}earning (\textbf{\methnametrim}), a self-supervised learning framework that supplementarily captures local graph information with masking-based modeling compared with vanilla contrastive learning. 
Extensive experiments validate the superiority of \methname against state-of-the-art methods, demonstrating its promise as a comprehensive graph representation learner.

\end{abstract}

\vspace{-2pt}
\section{Introduction}
\vspace{-3pt}
\label{sec:intro}




Graph data, with its rich topological features, has gained prominence across diverse scenarios such as social networks \cite{tang2008arnetminer} and biochemical reactions \cite{rong2020-selfsupervised}. Graph representation learning (GRL) methods handle the complexities of graph data by encoding these data into low-dimensional embeddings, effectively retaining both topological and feature information. These techniques have found their way into areas like drug discovery \cite{gaudelet2021utilizing} and social network analysis \cite{fan2019graph}.
Meanwhile, self-supervised learning (SSL), a method that leverages unlabeled data for feature learning, has gained growing attention across a wide range of fields such as computer vision \cite{misra2020self, park2023self}. This growing interest is largely driven by scenarios where task-specific labels are either scarce or expensive to acquire. For example, in biology and chemistry, obtaining labels through wet-lab experiments frequently requires substantial resources and time \cite{zitnik2018prioritizing}.

As a leading self-supervised representation learning method, Contrastive learning (CL) has been demonstrating to be highly effective in natural language processing \cite{devlin2018bert} and computer vision \cite{chen2020simple}. The key idea of contrastive learning is to maximize the agreement between views generated from the same data instance, while minimizing the agreement between those from different instances. 
Given its potential in exploiting data relationships, recent efforts \cite{you2020-graph,zhu2021graph,chu2021cuco} have been devoted to advancing contrastive learning to obtain general graph representations. 
For example, GraphCL \cite{you2020-graph} builds a graph contrastive learning framework with augmentations to learn robust graph representations. 
GCA \cite{zhu2021graph} optimizes contrastive objective by employing adaptive data augmentation on graph topology and attributes.
Cuco \cite{chu2021cuco} combines curriculum learning with contrastive learning so that negative samples can be trained in a human-learning manner.
However, representations obtained by graph contrastive learning are far from general representations. One primary reason lies in the method's approach to differentiating between two similar graph views, where identifying the global patterns is usually more important than capturing local structures. The contrastive objective leaves graph contrastive learning overemphasized on global pattern and insensitive to local structures that also contain crucial graph information.

\begin{figure*}[tb]
\centering
\includegraphics[width=0.9\textwidth]{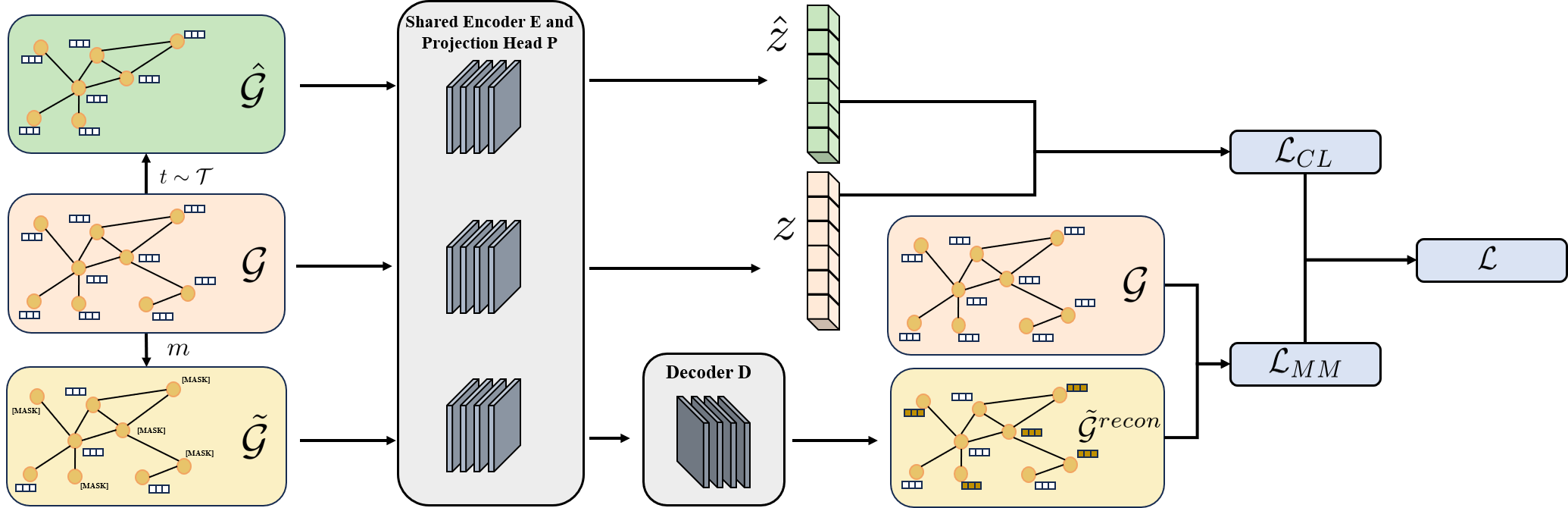}
\vspace{-5pt}
\caption{The framework of \methnametrim. \methname enhances graph contrastive learning to learn more informative representations containing local graph patterns by leveraging masking-based modeling objective.}
\vspace{-13pt}
\label{fig:framework}
\end{figure*}

In this paper, we address the aforementioned challenge by adopting principles from the masking-based modeling objective that have been successful in computer vision \cite{park2023self,he2022masked}. Such methods are dedicated to learning representations from masked images with the restoration goal, emphasizing the capture of local details. Motivated by these successes, we propose \underline{Local}-aware \underline{G}raph \underline{C}ontrastive \underline{L}earning (\textbf{\methnametrim}), a novel self-supervised learning framework that achieves a comprehensive graph representation by understanding both global patterns and local structures.
Our contributions can be summarized as follows:

\begin{itemize}
\vspace{-5pt}
\item We introduce \methnametrim, a self-supervised graph representation framework, addressing inadequate focus on local information in current graph contrastive learning techniques. By incorporating a masking-based modeling objective, our method grasps both global and local structures of graphs, offering more informative graph representations.
\vspace{-5pt}
\item We empirically demonstrate that \methname is an informative self-supervised graph representation learner by showing it outperforms baseline models through extensive experiments.
\end{itemize}

\vspace{-2pt}
\section{Preliminary}
\vspace{-3pt}
Let $\mathcal{G} = \{\mathcal{V}, \mathcal{E}\}$ denote an undirected graph, where $\mathcal{V} = \{v_1, v_2, \dots, v_{N}\}$ is the node set with $\mathbf{X}\in \mathbb{R}^{N\times d}$ as the feature matrix, $\mathcal{E} \subseteq \mathcal{V} \times \mathcal{V}$ is the edge set with $\mathbf{A} \in \{0,1\}^{N\times N}$ as the adjacency matrix. Our objective is to learn an encoder $\mathbf{E}(\mathbf{X}, \mathbf{A})$ which uses graph features and structures as input and produces graph embeddings in low dimensionality. These embeddings can be used in downstream tasks such as graph classification and transfer learning.

\vspace{-2pt}
\section{The LocalGCL Framework}
\label{sec:meth}
\vspace{-3pt}

In the framework of \methname shown in Fig. \ref{fig:framework}, we utilize the training objectives of contrastive learning in conjunction with masking-based modeling through a multi-task learning approach. The training pipeline of \methname includes data processing, encoding, and sub-task training.

\vspace{-2pt}
\subsection{Data Processing}
\vspace{-5pt}

In the data processing phase, \methname generates an augmented view and a masked view of graph $\mathcal{G}$. 
At each iteration, it stochastically samples an augmentation function $t \sim \mathcal{T}$, where $\mathcal{T}$ is a set of all possible augmentation functions. In this work, we adopt four data augmentation techniques including node dropout, edge perturbation, attribute masking and subgraph \cite{you2020-graph}. Then, an augmented view of graph  $\mathcal{G}$ is generated by $t$, denoted as $\hat{\mathcal{G}} = t(\mathcal{G})$. 

Meanwhile, \methname applies masking on $\mathcal{G}$ to generate corresponding masked view $\tilde{\mathcal{G}}$. Masking can be applied on structures $\mathbf{A}$ or on features $\mathbf{X}$. Though many works focus on structure masking \cite{kipf2016variational,pan2018adversarially}, our approach adopts feature masking since features can provide a local view of what each node represents. Formally, our model generates a masked view using the feature masking function $m$, denoted as $\tilde{\mathcal{G}} = m(\mathcal{G})$.

\vspace{-2pt}
\subsection{Encoding}
\vspace{-5pt}

During encoding, different graph views from previous phase are fed into both the encoder and projection head to obtain embeddings in the projection space. Within our framework, the encoder $\mathbf{E}$ which comprises several layers of graph neural networks (GNNs) is shared across various tasks. 
The projection head $\mathbf{P}$ is trained to transform the encoder's output into more robust representations that are invariant to augmentations. The backbone for the encoder can be any type of GNNs, such as GIN \cite{xu2019-how}, GCN \cite{kipf2016semi} or GAT \cite{velivckovic2018graph}, while a multi-layer perceptron (MLP) is employed as the structure of projection head as advocated by \cite{chen2020simple}. 
Specifically, embeddings $\bm{z} = \mathbf{P}(\mathbf{E}(\mathbf{X}, \mathbf{A}))$, $\hat{\bm{z}} = \mathbf{P}(\mathbf{E}(\hat{\mathbf{X}}, \hat{\mathbf{A}}))$ and $\tilde{\bm{z}} = \mathbf{P}(\mathbf{E}(\tilde{\mathbf{X}}, \tilde{\mathbf{A}}))$ are generated by encoder $\mathbf{E}$ and projection head $\mathbf{P}$. 

\vspace{-2pt}
\subsection{Sub-Task Training}
\vspace{-5pt}

After obtaining embeddings from projection space, a contrastive objective and a masking-based objective are employed to optimize the model parameters. 
For the contrastive objective, the aim is to distinguish embeddings of the original graph and its augmented variant from other negative samples. Embeddings from varied augmentations of the same original graph are pulled closer while those from different graphs are pushed away through minimizing normalized temperature-scaled cross entropy loss (NT-Xent) \cite{sohn2016-improved,wu2018-unsupervised}, denoted as:


\vspace{-8pt}
\begin{equation}
    \mathcal{L}_{CL} = -\sum^{N}_{i=1}\log \frac{\exp(\textrm{SIM}(\bm{z}_{i},\hat{\bm{z}_{i}})/\tau)}{\sum^{N}_{j=1,j\ne i}\exp(\textrm{SIM}(\bm{z}_{i},\hat{\bm{z}_{j}})/\tau)},
\label{eq:lcl}
\end{equation}
\vspace{-5pt}

where $\textrm{SIM}(\cdot)$ denotes the cosine similarity function, with $\bm{z}_{*}$ and $\hat{\bm{z}_{*}}$ as the embeddings of $G_{*}$ and $\hat{G_{*}}$ in the projection space, respectively.

For the masking-based objective, a decoder $\mathbf{D}$ reconstruct the embeddings $\tilde{\bm{z}}$ back to the graph space, denoted by $\tilde{\mathcal{G}}^{recon} = \mathbf{D}(\tilde{\bm{z}})$. 
The objective is to maximize the agreement between the reconstructed graph $\tilde{\mathcal{G}}^{recon}$ and the original graph $\mathcal{G}$. 
Specially, the mean square error is applied upon these graph views, formally written as:

\vspace{-8pt}
\begin{equation}
    \mathcal{L}_{MM}=\frac{1}{N} \sum_{i=1}^{N} ||\mathbf{X}_{i} - \mathbf{\tilde{X}}_{i}^{recon}||_F^2,
\label{eq:lmm}
\end{equation}
\vspace{-5pt}

where $\mathbf{X}_{i}$ and $\mathbf{X}_{i}^{recon}$ are feature matrices for $\mathcal{G}_{i}$ and $\tilde{\mathcal{G}}_{i}^{recon}$ respectively.

These two objectives are balanced through a linear combination:

\vspace{-8pt}
\begin{equation}
    \mathcal{L}=(1-\lambda) \mathcal{L}_{CL} + \lambda \mathcal{L}_{MM},
\label{eq:loss}
\end{equation}
\vspace{-5pt}

where $\lambda \in [0,1]$ serves as a hyperparameter that controls the relative importance of the contrastive and masking-based objectives within the total loss. 
The training algorithm is summarized in Alg. \ref{alg:training}.

\begin{algorithm}[tb]
	\caption{The \methname training algorithm}
	\label{alg:training}
	\textbf{Input:} $\mathcal{B}$: input batch of $N$ graphs where $\mathcal{G}_{i} = \{\mathcal{V}_{i}, \mathcal{E}_{i}\}$ is the i-th graph in this batch with $\mathbf{X}_{i}$ as feature matrix and $\mathbf{A}_{i}$ as adjacency matrix.
        $\mathbf{M}(\cdot) = \{ \mathbf{E}(\cdot), \mathbf{P}(\cdot), \mathbf{D}(\cdot) \}$: the training model with $\mathbf{E}(\cdot)$ as encoder, $\mathbf{P}(\cdot)$ as projection head and $\mathbf{D}(\cdot)$ as decoder.\\
	\begin{algorithmic}[1]
        \vspace{-10pt}
        \FOR{ each $\mathcal{G}_{i}$ in $\mathcal{B}$}
            \STATE Sample an augmentation function $t \sim \mathcal{T}$.
            \STATE Generate an augmented view $\hat{\mathcal{G}_{i}} = t(\mathcal{G}_{i})$ and a masked view $\tilde{\mathcal{G}_{i}} = m(\mathcal{G}_{i})$.
            \STATE Obtain embeddings $\bm{z}_{i}$, $\hat{\bm{z}_{i}}$ and $\tilde{\bm{z}_{i}}$ of $\mathcal{G}_{i}$, $\hat{\mathcal{G}_{i}}$ and $\tilde{\mathcal{G}_{i}}$ respectively using the encoder $\mathbf{E}(\cdot)$ and projection head $\mathbf{P}(\cdot)$.
            \STATE Obtain the reconstructed graph $\tilde{\mathcal{G}}_{i}^{recon}$ using the decoder $\mathbf{D}(\cdot)$.
        \ENDFOR

        \STATE Compute the contrastive objective $\mathcal{L}_{CL}$ with Eq. \ref{eq:lcl}.
        \STATE Compute the masking-based objective $\mathcal{L}_{MM}$ with Eq. \ref{eq:lmm}.
        \STATE Determine $\lambda$ based on the training progress.
        \STATE Update parameters of $\mathbf{M}$ by minimizing $\mathcal{L}$ with Eq. \ref{eq:loss}.
	\end{algorithmic}
\end{algorithm}



In \methnametrim, a crucial parameter in training progress is \( \lambda \), which defines the proportional significance of two distinct loss functions. 
When applied a static strategy, \( \lambda \) serves as a hyper parameter that should be optimized according to specific datasets. However, the importance of contrastive objective and masking-based objective is dependent on the training process. During the initial training stages, contrastive learning provides a broad global viewpoint. As training progresses,  a shift towards emphasizing local features emerges for refinement, thereby amplifying the masking-based modeling objective. Thus, in this work, we adaptively modulate \( \lambda \) using incremental dynamic strategy. This strategy assigns a relatively low value to \(\lambda\) so that the model can first capture a global perspective of graphs. It shifts model's attention to masking-based objective by gradually increasing the value of \( \lambda \). We empirically validate the effectiveness of incremental dynamic strategy in Sec. \ref{sec:ablation}.
This dynamic weighting strategy combines the strengths of both learning methods, enabling the model to address local features and global architectures to learn more informative graph representations.


\vspace{-2pt}
\section{Experiments}
\vspace{-3pt}
\label{sec:exp}

In this section, we demonstrate that \methname is an informative self-supervised framework for graph representation learning by conducting extensive experiments on unsupervised graph classification task (Sec. \ref{sec:classification}) and transfer learning task (Sec. \ref{sec:transfer}). In addition, an ablation study is performed to investigate the significance of different strategies on \( \lambda \) in Eq. \ref{eq:loss} (Sec. \ref{sec:ablation}).

\begin{table*}[tb]
\renewcommand{\arraystretch}{0.7} 
  \centering
  \vspace{-10pt}
  \caption{Experiment results for unsupervised graph classification in accuracy(\%) with mean and std. We highlight the best results in \textbf{bold} for each dataset. The reported results of baselines are from previous papers if available.}
    \vspace{-10pt}
    \begin{tabular}{c|ccc|ccc}
    \toprule
    Datasets & IMDB-B & IMDB-M & COLLAB & MUTAG & PROTEINS & NCI1 \\
    \midrule
    WL    & 72.30±3.44 & 46.95±0.46 & -     & 80.72±3.00 & 72.92±0.56 & 80.31±0.46 \\
    DGK   & 66.96±0.56 & 44.55±0.52 & -     & 87.44±2.72 & 73.30±0.82 & 80.31±0.46 \\
    GraphMAE & 75.52±0.66 & 51.63±0.52 & 80.32±0.46 & 88.19±1.26 & \textbf{75.30±0.39} & \textbf{80.40±0.30} \\
    GraphCL & 71.14±0.44 & 48.58±0.67 & 71.36±1.15 & 86.80±1.34 & 74.39±0.45 & 77.87±0.41 \\
    \midrule
    ours  & \textbf{75.64±0.25} & \textbf{52.26±1.07} & \textbf{82.68±0.74} & \textbf{89.26±1.56} & 74.72±1.08 & 78.07±0.52 \\
    \bottomrule
    \end{tabular}%
  \label{tab:results}%
\end{table*}%

\begin{figure*}[tb]
    \renewcommand{\arraystretch}{0.7} 
    \setlength{\tabcolsep}{3pt}
    \centering
    \begin{minipage}{0.4\textwidth} 
        \vspace{-5pt}
        \captionof{table}{Experiment results for transfer leaning in ROC-AUC(\%) with mean and std. The best results are in \textbf{bold} for each dataset.}
        \vspace{-9pt}
        \begin{adjustbox}{valign=c}
            \scalebox{0.85}{
                \begin{tabular}{c|cccc}
                    \toprule
                    Datasets & BACE  & HIV   & MUV   & ClinTox \\
                    \midrule
                    No-pretrain & 70.0±2.5 & 75.4±1.5 & 71.7±2.3 & 58.2±2.8 \\
                    \midrule
                    AttrMasking & 79.3±1.6 & 77.2±1.1 & 74.7±1.4 & 71.8±4.1 \\
                    Infomax & 75.9±1.6 & 76.0±0.7 & 75.3±2.5 & 69.9±3.0 \\
                    GraphCL & 75.4±1.4 & \textbf{78.5±1.2} & 69.8±2.7 & 76.0±2.7 \\
                    GraphMAE & 83.1±0.9 & 77.2±1.0 & 76.3±2.4 & \textbf{82.3±1.2} \\
                    \midrule
                    LocalGCL(ours) & \textbf{83.3±1.2} & 77.6±2.5 & \textbf{76.5±1.7} & 77.8±3.3\\
                    \bottomrule
                \end{tabular}%
            }
        \end{adjustbox}
        \vspace{-9pt}
        \label{tab:transfer}
    \end{minipage}
    \hspace{0.2cm}
    \begin{minipage}{0.26\textwidth}  
        \centering
        \vspace{-5pt}
        
        \begin{adjustbox}{valign=c}
            \includegraphics[width=\linewidth]{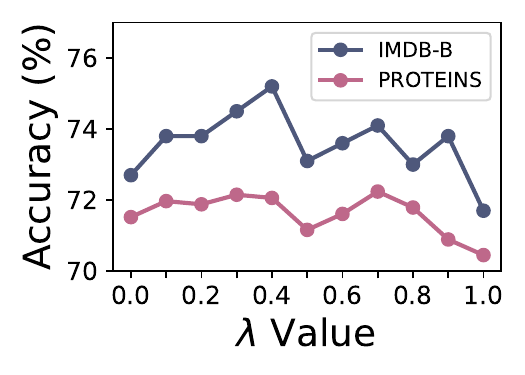}
        \end{adjustbox}
        \vspace{-10pt}
        \caption{Experiment results for different static $\lambda$ strategies.}
        \vspace{-16pt}
        \label{fig:static}
    \end{minipage}%
    \hspace{0.2cm}
    \begin{minipage}{0.28\textwidth}  
        \centering
        \vspace{-3pt}
        \begin{adjustbox}{valign=c}
            \includegraphics[width=\linewidth]{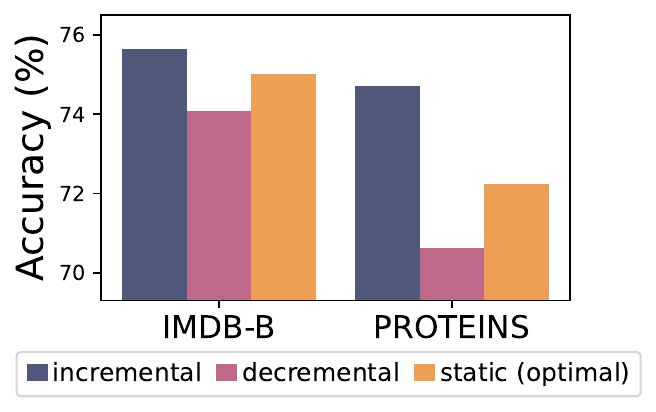}
        \end{adjustbox}
        \vspace{-10pt}
        \caption{Comparison of dynamic and static $\lambda$ strategies.}
        \vspace{-16pt}
        \label{fig:lambda}
    \end{minipage}

\end{figure*}


\vspace{-2pt}
\subsection{Performance Comparison and Analysis}
\vspace{-5pt}
\subsubsection{Unsupervised Graph Classification}
\label{sec:classification}
\vspace{-5pt}


\textbf{Settings.} For unsupervised graph classification task, we use several benchmark graph datasets from varied domains including bioinformatics datasets like MUTAG, NCI1 and PROTEINS containing molecular and protein data useful for drug design as well as social networks datasets like IMDB-BINARY, IMDB-MULTI and COLLAB that highlight social network interactions and structures. 
During evaluation, we emphasize \methnametrim's ability to extract deep semantic information from graph data. We follow a generic evaluation method: graph-level representations are obtained using \methnametrim's encoder and subsequently classified using LIBSVM \cite{chang2011libsvm}, a renowned SVM classifier. The accuracy and standard deviation from the classifier give us clear performance metrics. Accuracy indicates effectiveness of representing original data's key semantics, while the standard deviation shows \methnametrim's consistency across repeated tests. Together, these metrics provide a comprehensive framework for evaluating the generalizability of graph representation models.

To evaluate the effectiveness of \methnametrim, we choose typical methods from both traditional graph kernels and self-supervised learning techniques for comparison. Specifically, we consider classic techniques like the Weisfeiler-Lehman subtree kernel (WL) \cite{shervashidze2011weisfeiler} and the Deep Graph Kernel (DGK) \cite{yanardag2015deep}, as well as baseline self-supervised methods such as masking-based modeling method GraphMAE \cite{hou2022-graphmae} and contrastive learning method GraphCL \cite{you2020-graph}.

\vspace{2pt}
\noindent\textbf{Results.} In the unsupervised graph classification experiments presented in Tab. \ref{tab:results}, \methname consistently demonstrates superior or competitive performance relative to the established baselines.
On the social network datasets, \methname  significantly stands out against other methodologies in terms of classification accuracy. Considering that these datasets capture the complex structures and relationships inherent in social networks, the superior performance achieved by \methname  highlights its capability of understanding these connections and the inherent complexities of social network data.
On the bioinformatics datasets, \methname surpasses several baselines, particularly outperforming GraphCL. However, when compared to GraphMAE, it shows a slight lag. 
This might be attributed to the significant influence of local details on molecular functionality in these datasets, which are rich in complex structural details and critical semantic information for bioinformatics and pharmaceutical research.
While our approach manages local information competently, GraphMAE appears to excel in exploiting certain aspects, signaling opportunities for further improvement for \methnametrim.


\vspace{-5pt}
\subsubsection{Transfer learning}
\label{sec:transfer}
\vspace{-5pt}

\textbf{Settings.} 
To assess the method's transferability, we conduct experiments on transfer learning for molecular property prediction following the setting of \cite{hu2019strategies}. The model is pretrained on ZINC15 dataset \cite{sterling2015zinc} and finetuned on different datasets in MoleculeNet \cite{wu2018moleculenet}. The datasets are split by scaffold to simulate real-world scenarios. 
The experiments are repeated for 10 times with different seeds and the results are compared in ROC-AUC scores with mean and standard deviation. Our approach is compared with AttrMasking \cite{hu2019strategies}, GraphCL \cite{you2020-graph} and GraphMAE\cite{hou2022-graphmae}.

\vspace{2pt}
\noindent\textbf{Results.} The results are shown in Tab. \ref{tab:transfer}. Notably, 
\methname achieves competitive results on all datasets and outperforms other SOTA methods on two datasets. 
This evidences not only the competitive edge of \methname but also its transferability and generalizability across diverse datasets.

\vspace{-2pt}
\subsection{Ablation Study}
\label{sec:ablation}
\vspace{-5pt}
The selection of the weight parameter \( \lambda \) is pivotal in the performance of \methnametrim. 
This study differentiates between static and dynamic \( \lambda \) strategies. For static strategies,  we examine a range of fixed \( \lambda \) values from 0 to 1 in increments of 0.1. While for dynamic strategies, we focus on the incremental and decremental approaches. 
The incremental dynamic strategy gradually increases \( \lambda \) during training process. Conversely, the decremental dynamic strategy reduces \( \lambda \) over time.
We test these strategies on IMDB-B and PROTEINS datasets for unsupervised graph classification following the settings of Sec. \ref{sec:classification}.

The performance of static strategies can be seen in Fig. \ref{fig:static}. 
On both datasets, the hybrid strategies outperform pure contrastive learning strategy (\( \lambda = 0 \)) and pure masking-based modeling strategy (\( \lambda = 1 \)). 
The optimal \( \lambda \) value is different on different datasets. 
For the IMDB-B dataset, accuracy peaks around \(\lambda = 0.4\) for the IMDB-B dataset as it increases from 0. Following this peak, accuracy exhibits fluctuations with an overall decline up to \( \lambda = 1 \). 
For the PROTEINS dataset, the optimal \( \lambda \) value is relatively higher around \(0.7\). 
This suggests that different datasets might emphasize differently on the global and local information in graphs.
Thus, a proper selection of \(\lambda \) is needed in finetunig the performance of our approach when static strategy is applied.


As depicted in Fig. \ref{fig:lambda}, we compare incremental and decremental strategies with best results from static strategies to investigate the effectiveness of dynamic strategies. The incremental dynamic strategy yields superior performance on both datasets. This suggests the model benefits from an initial emphasis on global features, transitioning to local features as training progresses. Conversely, the decremental dynamic strategy, which prioritizes local features initially, underperforms, underscoring its suboptimality. 



In summary, dynamic strategies, especially the incremental approach that initially emphasizes global features, demonstrate enhanced effectiveness across datasets.

\vspace{-2pt}
\section{Conclusion}
\vspace{-3pt}
\label{sec:concl}

In this paper, we address the limitation of GCL in generating less informative graph representations due to its disproportionate focus on global patterns over local structures.
Thus, we propose \methnametrim, a framework that refines contrastive learning through masking-based modeling objectives and dynamically harmonizes balance of these objectives.
\methname offers more informative representations that effectively exploit both global and local information in graphs. 
Extensive experiments across diverse datasets validate the effectiveness of \methnametrim.

\vfill\pagebreak



\bibliographystyle{IEEEbib}
\bibliography{ICASSP2024}

\end{document}